\documentclass[]{interact}
\usepackage[utf8]{inputenc}
\usepackage[bottom]{footmisc}
\usepackage[caption=true]{subfig}

\usepackage{natbib}

\theoremstyle{plain}

\theoremstyle{definition}

\theoremstyle{remark}

\usepackage{tikz}
\usetikzlibrary{positioning}
\usepackage{graphicx}
\usepackage{import}

\usepackage{fancyhdr}
\usepackage{lipsum} 

\fancypagestyle{plain}{
  \fancyhf{}
  \fancyfoot[C]{\thepage}%
}
\fancypagestyle{firstpage}{
  \fancyhf{}
  \fancyfoot[L]{This is a Preprint of an article accepted for publication by Inderscience in International Journal of Computational Vision and Robotics on September 2018. DOI: https://dx.doi.org/10.1504/IJCVR.2019.102286}%
}
\begin{document}

\title{\Large{Combination of Domain Knowledge and Deep Learning for Sentiment Analysis of Short and Informal Messages on Social Media}}
\maketitle
\textbf{Khuong Vo, Tri Nguyen, Dang Pham, Mao Nguyen, Minh Truong} \\
YouNet Group, Ho Chi Minh City, Vietnam\\
\{khuongva, trind, dangpnh, maonx, minhta\}@younetgroup.com\\

\textbf{Trung Mai, Tho Quan} \\
Ho Chi Minh City University of Technology, Vietnam National University - HCMC, Ho Chi Minh City, Vietnam\\
\{mdtrung, qttho\}@hcmut.edu.vn\\
\thispagestyle{firstpage}
\pagebreak

\begin{abstract}
\textit{Sentiment analysis} has been emerging recently as one of the major \textit{natural language processing} (NLP) tasks in many applications. Especially, as \textit{social media channels} (e.g. social networks or forums) have become significant sources for brands to observe user opinions about their products, this task is thus increasingly crucial. However, when applied with real data obtained from social media, we notice that there is a high volume of \textit{short and informal messages} posted by users on those channels. This kind of data makes the existing works suffer from many difficulties to handle, especially ones using deep learning approaches. In this paper, we propose an approach to handle this problem. This work is extended from our previous work, in which we proposed to combine the typical deep learning technique of Convolutional Neural Networks with domain knowledge. The combination is used for acquiring additional training data augmentation and a more reasonable loss function. In this work, we further improve our architecture by various substantial enhancements, including negation-based data augmentation, transfer learning for word embeddings, the combination of word-level embeddings and character-level embeddings, and using multitask learning technique for attaching domain knowledge rules in the learning process. Those enhancements, specifically aiming to handle short and informal messages, help us to enjoy significant improvement in performance once experimenting on real datasets.
\end{abstract}

\pagebreak

\pagebreak

\section{Introduction}
\subsection{Sentiment analysis}
\textit{Opinion} was defined by Oxford Dictionary as the feeling or the thought of someone about something and these thoughts are not necessarily the truth. Therefore, opinion is always an important reference for making decisions of individuals and organizations. The term \textit{opinion mining/sentiment analysis} \citep{liu2012sentiment} thus has been coined and is developed rapidly and attracting much attention in research communities. With the recent advancement of the information and communications technology (ICT) era, sentiment analysis has been applied in various domains, such as medical domain \citep{sarker2011outcome, costumero2014approach,biyani2014identifying}, stock market \citep{yu2013using}, news analysis \citep{hagenau2013automated} and political debates \citep{walker2012your}.

Research on this topic was conducted at different levels: term level \citep{ding2010resolving}, phrase level \citep{kim2009discovering}, sentence level \citep{kim2009discovering} and document level \citep{turney2002thumbs,pang2002thumbs}. In terms of methodologies, approaches related to this problem can be summarized as follows.

\begin{itemize}
    \item \emph{Lexicon approach}: Sentiments terms are used a lot in sentiment analysis. There are positive terms and negative terms. Additionally, there are also opinion phrases or idioms, which can be grouped into Opinion Lexicon \citep{Taboada2011}. The dictionary-based method by Minging and Kim \citep{hu2004mining,kim2004determining} shows strategies using a dictionary for identifying sentiment terms.
    \item \emph{Corpus-based methods}: This approach is based on syntax and pattern analysis to find sentiment words in a big dataset \citep{hatzivassiloglou1997predicting}.
\end{itemize}
 
Recently, with the introduction of \emph{TreeBank}, especially \emph{Stanford Sentiment Treebank} \citep{socher2013recursive}, sentiment analysis using deep learning becomes an emerging trend in the field. \emph{Recursive Neural Tensor Networks} (RNTNs) were applied to the treebank and produced high performance \citep{socher2013recursive}. Formerly, the compositionality idea related to neural networks has been discussed by Hinton \citep{hinton1990mapping}, and the idea of feeding a neural network with inputs through multiple-way interactions, parameterized by a tensor have been proposed for relation classification \citep{jenatton2012latent}. Along with the treebank, the famous \emph{Stanford CoreNLP} tool \citep{manning2014stanford} is used widely by the community for sentiment tasks. Besides, the convolution-based method continues to be developed for sentiment analysis on sentences \citep{kim2014convolutional, zhang2015text}. To store occurrence order relationship between features, recurrent neural networks such as \emph{Long Short-Term Memory} networks (LSTMs) were used in combination with \textit{Convolutional Neural Networks} (CNNs) to perform sentiment analysis for short text \citep{wang2016combination}. Most recently, a combined architecture using deep learning for sentiment analysis has been proposed in \citep{nguyen2017deep}.

\subsection{Sentiment analysis for short and informal messages on social media}
Nowadays, through the Internet, \textit{social media} has emerged as an efficient channel to handle the social crisis \citep{middleton2014real, osborne2014real}. In the \textit{social networks} of \emph{Facebook} and \emph{Twitter} or electronic newspapers, the information is updated continuously from the user as a streamline of \textit{feed}. In such a dynamic environment, opinion detection of users may be very useful in many aspects \citep{maynard2014should}. For instance, it can help brands to quickly counter-attack social crisis related to their products or services \citep{conover2011political,kass2013social,zhao2015enquiring}.

In those social media channels, messages posted by users tend to be \textit{short}, usually spanning one sentence or less, and the language used is \textit{informal}. Working with short and informal textual messages involves confronting many challenges that make sentiment analysis difficult. Those messages commonly contain many intentional and unintentional misspellings, elongated words, slang terms and shortened forms of words. Moreover, the sentiment contexts are often lacked when conveyed in such messages.

Various attempts have been reported to address this issue, such as the classical lexicon-based approach \citep{thelwall2010sentiment,medhat2014sentiment}, leveraging a variety of surface form, semantic, and sentiment features for statistical text classification approach \citep{Kiritchenko2014}, or using state-of-the-art CNNs that exploit from character- to sentence-level information \citep{Santos2014}. However, the application of deep learning techniques for the problem of short and informal textual messages have suffered from the following drawbacks.

\begin{itemize}
    \item Negation form (such as \textit{``I don't like this mobile model''}) is often used in short and informal messages. To express that the sentiment meaning of the sentence is inverted by the use of such negation phrase is quite simple with a rule-based approach. However, a \textit{deep neural network} (DNN) hardly learns such rule, mostly due to the insufficient training data when applied in real applications. To let the DNNs learn and adapt to the negation form, the training data should cover all of the cases that negation adverbs/phrases are used, which is difficult to implement.
    \item Continuously training the pre-trained word embeddings for the sentiment task, to better reflect the polarity difference among sentiment terms, is not ideal for this kind of tasks as annotated data is often noisy.
    \item In informal messages, chances of misspelling and other typos are relatively higher than those in formal documents. Users even do it on purpose to emphasize their feelings. For example, users may stress \textit{``Nooooooo!''} rather than just simply saying \textit{``No!''} to indicate their highly disappointed attitude. Training a DNN to well adapt with such misspelling cases is not a trivial task.
    \item Since the message length is short, the sentiment context is hardly captured merely by the deep architecture alone. Integrating domain knowledge, usually represented by \textit{semantic rules}, into the training process of a DNN would become highly crucial in this case.
\end{itemize}
In this paper, we deal with those drawbacks by extending our works in \citep{vo2017combination}, in which some ideas of combining domain knowledge with deep learning have been drawn. In this work, we introduce a new architecture, known as SAINT, where the following enhancements are made.
\begin{itemize}
    \item We adapt the previous technique of term-based data augmentation to \textit{enriching the training data with negation cases}.
    \item We \textit{transfer the stable word embeddings} that were learned in the previous variable-length message sentiment task to the short-length message task to improve the performance of the embedding layer.
    \item We suggest to combine \textit{word-level embeddings with character-level embeddings}. This approach adds more contextual information to each word, allowing various misspelling cases from the same original words to be embedded in extra similar vectors.
    \item We also introduce \textit{multi-task learning} technique which allows the syntactic rules, which are efficient to handle short messages, to be captured and trained together with the document contents.
\end{itemize}

Our enhancements have enjoyed remarkable improvements when applied with real datasets collected from social media channels. The rest of our paper is organized as follows. Section \ref{sec:preliminaries} presents the background knowledge required to understand this study. Section \ref{sec:rules} discusses some syntactic rules, which are efficient for handling short texts. Section \ref{sec:proposed-approach} presents our proposed architecture. In this section, our previous work is firstly given, followed by a new improved one. In Section \ref{sec:experiments}, experimental results are discussed. Finally, Section \ref{sec:conclusion} concludes the paper.

\section{Preliminaries} \label{sec:preliminaries}

\subsection{Word embeddings}

\begin{figure}[h]
\centering
\includegraphics[scale=0.45]{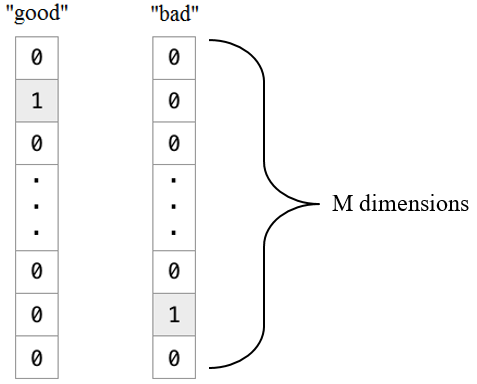}
\caption{Representing words by one-hot vector.} \label{fig:repr-one-hot}
\end{figure}

Word embedding basic mode is often used to perform weighting vectors. Basically, this is a weight vector, for example, 1-of-$N$ (one-hot vector), used to encode the word in a dictionary of $M$ words into a vector of length $M$. As presented in Figure \ref{fig:repr-one-hot} is a one-hot vector representing two words in a dictionary. However, using that simple representation, we can not evaluate the similarity between words since the distance between any two vectors are always the same (e.g. cosine distance). Moreover, the dimension of the vector space is huge when applied to the real dictionary.

\begin{figure}[h]
\centering
\includegraphics[scale=0.45]{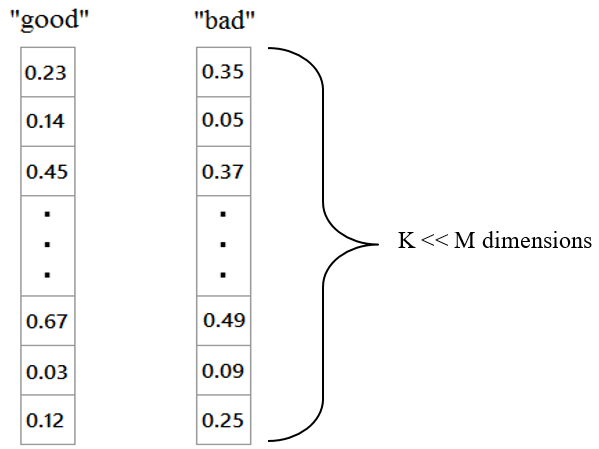}
\caption{Vectors representing relationships between words.} \label{fig:vector-words-relation}
\end{figure}

Word embedding model using \textit{Word2vec} technique \citep{mikolov2013distributed} represents the form of a distribution relationship of a word in a dictionary with the rest (known as \textit{Distributed Representation}). In this model, words are embedded in a continuous vector space where semantically similar words are represented in adjacent points. This is based on the idea that words sharing semantic meaning are in the same contexts. As presented in Figure \ref{fig:vector-words-relation}, each word now is represented as a $K-$dimensional vector, where $K<<M$, and each element of the vector is represented by new learning value.

\begin{figure}[h]
    \centering
    \includegraphics[scale=0.5]{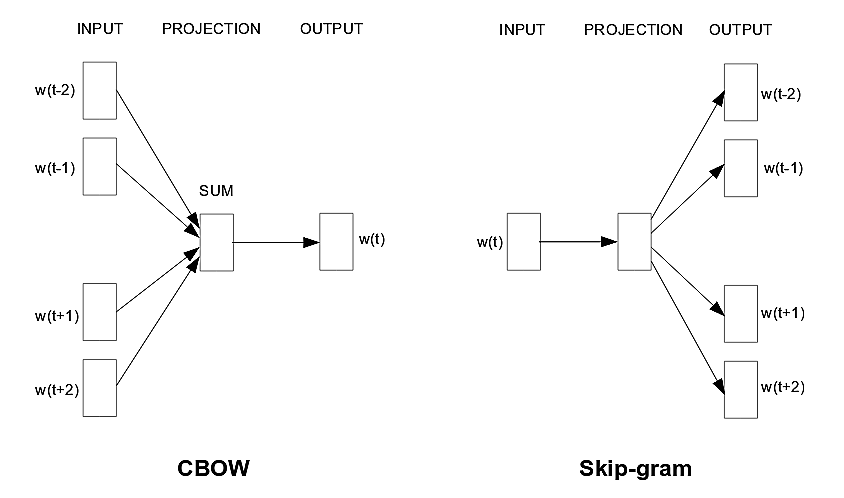}
    \caption{CBOW and Skip-gram \citep{mikolov2013efficient}}
    \label{fig:cbow-skipgram}
\end{figure}

Word2vec comes in two techniques, the \textit{Continuous Bag-of-Words} model (CBOW) and the \textit{Skip-Gram} model. Both of them come up with the same architecture of shallow three-layer neural networks (Figure \ref{fig:cbow-skipgram}), which the first layer's weights play as embedding vectors. In this network, the dimension of the second layer ($K$) is relatively lower than the input layer ($M$), which results in a lower dimension of the embedding vectors and makes computation more efficient. Despite sharing the same architecture, there are differences in input and output layers between CBOW and Skip-gram: CBOW model predicts a target word from the given context words, while the Skip-gram model predicts the context words from a given word.

\subsection{Convolutional Neural Networks for sentiment analysis}
\begin{figure}[h]
\centering
\includegraphics[scale=0.32]{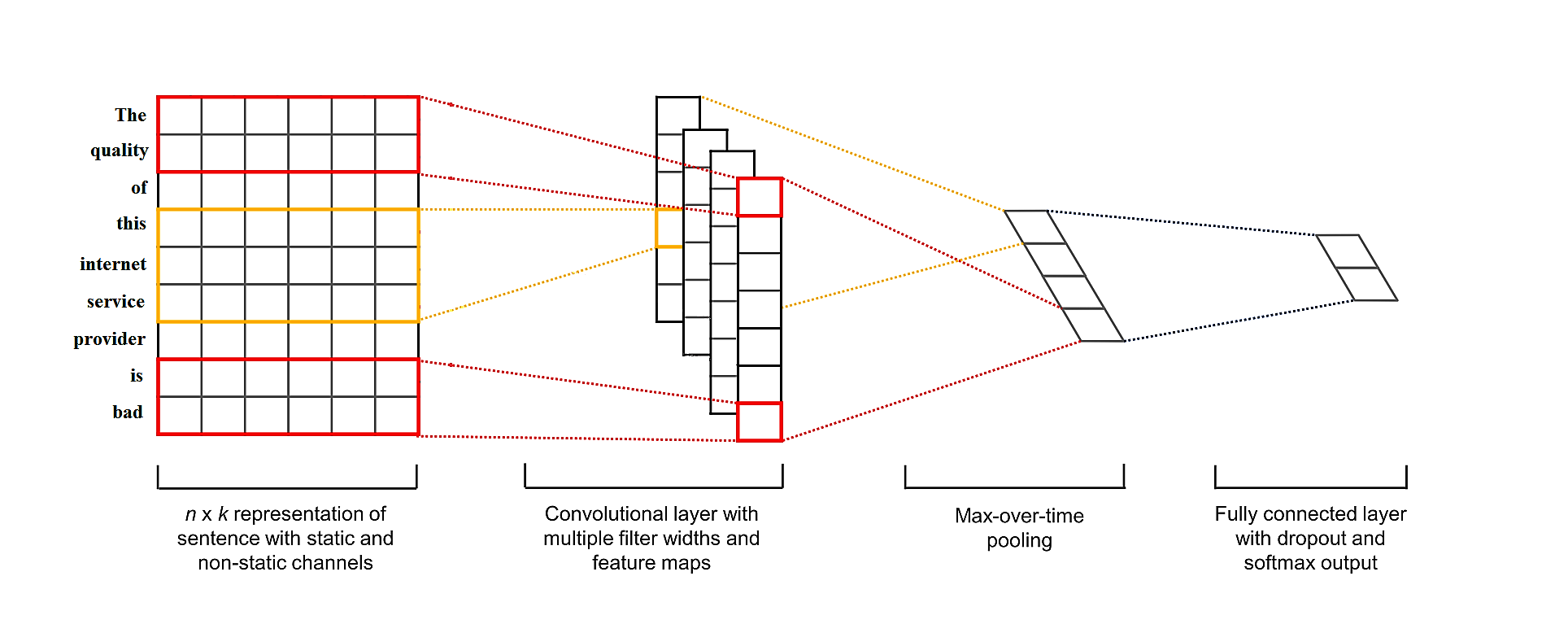}
\caption{Using CNNs for text processing \citep{kim2014convolutional}} \label{fig:cnn}
\end{figure}

\textit{Convolutional Neural Networks} (CNNs) are the popular deep learning models. Given in Figure \ref{fig:cnn} is the general architecture of such CNN system, typically used for NLP tasks. The first layer builds the vector from the words in the sentence. Input documents are transformed into a matrix, each row of which corresponds to a word in a sentence. For example, if we have a sentence with 10 words, each word was represented as a \textit{word-embedding} vector of 100 dimensions, the matrix has the size of $10 \times 100$. This is similar to an image with $10 \times 100$ pixels. The next layer performs \textit{convolution} on these vectors with different filter sets and then \textit{max-pooling} is performed for the set of filtered features to retain the most important features. Then, these features are passed to a fully connected layer with the softmax function to produce the final probability output. \textit{Dropout} \citep{srivastava2014dropout} technique is used to prevent overfitting.

In \citep{kim2014convolutional}, basic steps of using a CNN in sentiment analysis was detailed in the process by which one feature is extracted from one filter as follows.

Given a sentence with $n$ words, let $x_i \in R^k$ be a $k$-dimensional word vector corresponding to the $i$-th word in the sentence. The sentence can be represented as:
\textcolor{black}{$$x_{1:n} = x_1 + x_2 + ... + x_n$$}
Here, ``$+$'' denotes vector concatenation. Generally, \textcolor{black}{$x_{i:j}$ represents the word vector from index $i$ to $j$ ($j\geq i$)}. A convolution operator with filter $w \in R^{h \times k} $ for $h$ words produces the feature:
$$c_j = f(W.x_{i \colon i+h-1} + b),$$
Here, $b$ is the bias and $f$ is a non-linear function. By applying the filter on all windows of the sentence, we obtain the feature map:
$$c = [c_1, c_2, c_3, ..., c_{n -h + 1}]$$
The max-pooling is applied over the feature map and get the maximum value $\hat{c} = max\{c\}$ as the feature corresponding to this filter. Alternately, the max-pooling can be applied through local parts of the feature map to get local maximum values for this filter.

\subsection{Gated Recurrent Unit Networks}
\textit{Recurrent Neural Networks} (RNNs) are well known DNN architectures often used to handle sequential data. Basically, RNNs are arranged in a linear pattern, called \textit{state}, corresponding to a \textit{data entry} of input data. For example, to handle a text document, a state corresponds to a word in the text at timestep $t$. Each state received input including the corresponding data entry $x_t$ and the previous state $s_{t-1}$ to output the new state $h_t$. RNNs share the same parameters ($U$,$W$) across all steps.

In the 1990s, RNNs faced two major problems of \textit{Vanishing} and \textit{Exploding Gradients}. During the gradient backpropagation, the gradient signal can be multiplied many times as the number of timesteps by the weight matrix. If the weight is too small, the learned information is almost eliminated when the number of states becomes too large. Conversely, if the weight is large, this leads to the case known as the \textit{Exploding Gradients} when the gradient signal increasingly distracted during training, causing the process not converged.

\textit{Long Short-Term Memory} networks (LSTMs) \citep{hochreiter1997long} have similar structure to RNNs. However, they use a more complex way to compute the hidden state. LSTMs prevent the above problems through a gating mechanism in memory cells. They use gates which are the values from 0 to 1 to control how much of the information to let through at each timestep.

\textit{Gated Recurrent Unit} networks (GRUs) were proposed by \citep{cho2014learning} which is a variation of the LSTMs. GRUs also have \emph{gating mechanisms} that adjust the information flow inside the unit. However, they does not have an internal memory separated from its hidden state and combines the forget and input gates into a single update gate.
\begin{figure}[h]
\centering
\includegraphics[scale=0.35]{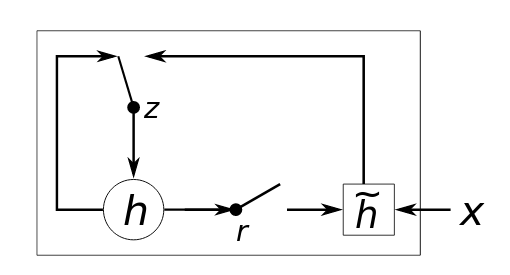}
\caption{Illustration of a GRU memory cell \citep{cho2014learning}} \label{fig:gru}
\end{figure}
A GRU memory cell, as illustrated in Figure \ref{fig:gru}, is updated at every timestep $t$ as the following equations:
\begin{align*}
    r_t &= \sigma (W_rx_t + U_rh_{t-1}) \\
    z_t &= \sigma (W_zx_t + U_zh_{t-1}) \\
    \hat{h}_t &= \tanh (Wx_t + U(r_t \odot h_{t-1}))\\
    h_t &= (1 - z_t)h_{t-1} + z_t\hat{h}_t
\end{align*}
where $\sigma$ denotes the logistic sigmoid function, $\odot$ is the elementwise multiplication, $r$ is the reset gate, $z$ is update gate and $\hat{h}_t$ denotes the candidate hidden layer.

Generally, both LSTMs and GRUs yield comparable performance \citep{jozefowicz2015empirical}; however, GRUs are computationally more efficient than LSTMs due to the less complex structure.

\begin{figure}[h]
\centering
\includegraphics[scale=0.55]{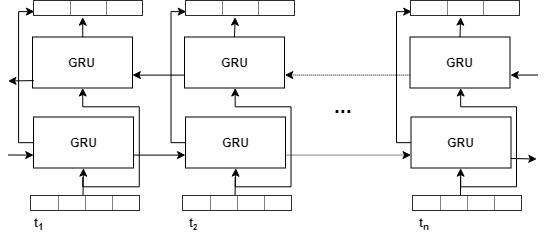}
\caption{Applying bidirectional strategy with GRUs} \label{fig:bi-gru}
\end{figure}

In addition, LSTMs/GRUs only focus on learning dependencies in one direction, with the assumption that the output at timestep $t$ only depends on previous timesteps. In this research, we deploy a concept as Bi-GRUs which borrow the idea from Bi-direction RNNs \citep{schuster1997bidirectional} to capture dependencies in both direction, as illustrated in Figure \ref{fig:bi-gru}.

\subsection{Transfer learning}
\textit{Transfer learning} is the process of using learned knowledge when solving a problem, which is usually trained weights, applied to solve another similar problem. When neural networks are able to learn more complex patterns, then the volume of data, especially labeled data, becomes much more crucial. In a few domains, adequately large enough datasets are published freely, e.g. \textit{ImageNet}. However, in the other domains, they are usually proprietary or expensive to obtain, or even not available. This makes transfer learning particularly important to deal with problems with small labeled datasets, by using a model trained on a free large similar dataset and then extracting some layer's weights to form a main model with a smaller dataset. 

In natural language processing domain, transfer learning was used as various forms. In \citep{Mou2016}, the authors investigated transfer learning for applying to:
\begin{itemize}
    \item Semantically similar tasks but different datasets.
    \item Semantically different tasks but sharing the same neural network \linebreak topology/architecture.
\end{itemize}
The conclusion of the effectiveness of using transfer learning in natural language processing is inconsistent. Although there are some semantic issues when applying transfer learning in NLP problems as compared to image processing \citep{Semwal2018, Mou2016}, word embeddings, which is a form of transfer learning, contributes on improving performance \citep{Mou2016}.

\subsection{Multi-task learning}
\begin{figure}[h]
    \centering
    \includegraphics[scale=0.4]{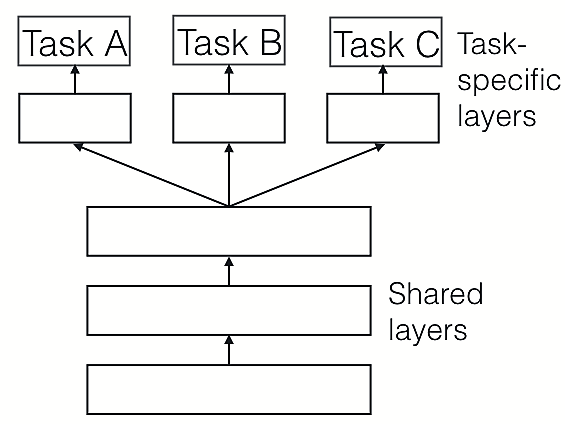}
    \caption{Multi-task learning architecture \citep{ruder2017overview}} 
    \label{fig:multi-taskling}
\end{figure}

\textit{Multi-task learning} \citep{Caruana1997, ruder2017overview} is an approach to learn a problem together with other related problems at the same time, using a shared representation. For example, one can do multi-label classification, where one sample is assigned multiple labels. Figure \ref{fig:multi-taskling} shows a network architecture learning multiple tasks at the same time.

Thus, multi-task learning mimics the way humans learn: We tend to learn multiple concepts at the same time and make connections to previous concepts when learning a new one. It also provides inductive bias from the auxiliary tasks, which causes the model to prefer hypotheses that explain more than one task, thereby improving the model's performance, i.e., it works as a form of regularization. Moreover, data from one task could be used to assist in the learning of the other tasks which has fewer data. For instance, in \citep{collobert2008unified}, the authors proposed a deep CNN for learning various NLP tasks at the same time, including part of speech (POS), chunking, named-entity recognition (NER), etc. and they reported an improvement in performance on these tasks.

\subsection{Using domain knowledge for sentiment analysis} \label{sec:using-domain-knowledge}
In general, when one performs sentiment analysis for a particular domain, the domain knowledge can be applied as shown in Figure \ref{fig:domain_knowledge}. Thus, a general system \citep{thanh2014sentiment} relies on a \textit{Sentiment Engine} to perform sentiment analysis on a user's comment expressing his opinion. This \textit{Sentiment Engine} operates based on a \textit{Knowledge Base} consisting of the following components:

\begin{figure}[h]
    \centering
    \includegraphics[scale=0.2]{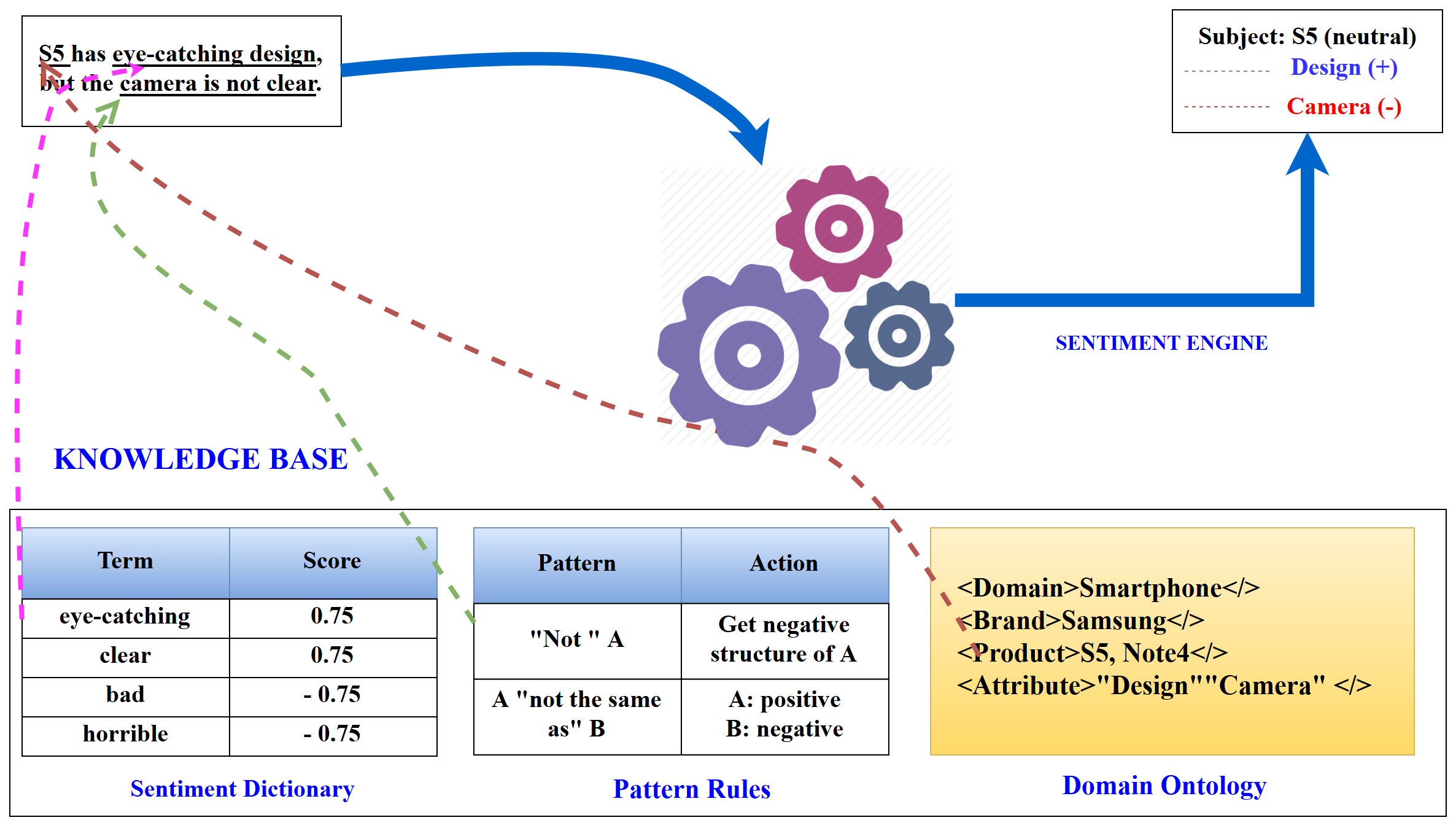}
    \caption{Applying domain knowledge for sentiment analysis}
    \label{fig:domain_knowledge}
\end{figure}

\begin{itemize}
    \item A sentiment dictionary, including the \textit{positive} and \textit{negative} sentiment terms. In particular, those sentiment words are assigned numerical scores indicating their \textit{sentiment levels}.
    \item Linguistic patterns are used to identify different phrase samples.
    \item A \textit{Sentiment Ontology} is for managing semantic relationships between sentiment terms and domain concepts. For more details on the Sentiment Ontology, please refer to \citep{thanh2014sentiment}.
\end{itemize}
Obviously, determining the sentiment scores for those sentiment terms is an important task to let such a system operate efficiently. In \citep{vo2017combination}, we did propose an approach to this problem.

\subsection{Data augmentation}
\textit{Data augmentation} is a technique commonly used in learning systems to increase the size of training datasets as well as to control generalization error for the learning model by creating different variations from the original data. For example, for image processing, one can re-size an image to generate different variants from this image. In our case, each sentiment term is associated with a \textit{sentiment score}, which is a value in the set \{-1, 0, 1\}. In \citep{vo2017combination}, we presented an approach which can infer sentiment scores from an annotated dataset. Then, from a training sample, we can generate term-based variants by replacing the sentiment terms in the original data by the other sentiment terms that \textit{have similar absolute scores}.

For example, let us consider an emotional sentence \textit{``Company A is better. Company B is horrible''}, with the object that needs to be analyzed being \textit{Company B}, the system first preprocesses the sentence as \textit{``Company A is better. \textbf{Target} is horrible.''} Obviously, this sentence is labeled as negative, w.r.t \textbf{Target}. In this example, we assume that words such as \textit{horrible, poor, terrible} have similar negative scores after our learning process. In addition, the words \textit{great} and \textit{amazing} have similar absolute values of opposite sign (i.e these words have positive scores). Thus, from this sample, we generate other augmented training samples as shown in Table \ref{tab:my_label}.

\begin{table}[h]
    \centering
    \caption{Examples of term-based training data augmentation}
    \begin{tabular}{|c|l | c|}
    \hline
        \# & Training Data & Label \\ \hline
        
        1 & \textit{Company A is better. Target is poor.} & Negative \\  \hline
        2 & \textit{Company A is better. Target is terrible} & Negative \\ \hline
        3 & \textit{Company A is worse. Target is great.} & Positive \\ \hline
        4 & \textit{Company A is worse. Target is amazing} & Positive \\ \hline
    \end{tabular}
    
    \label{tab:my_label}
\end{table}

In our learning system, the generation of augmented positive samples from the original negative samples is important, as this helps the system recognize that the word \textit{Company A} does not play any role in identifying emotions since it appears in both positive and negative samples. Conversely, sentiment orientation is determined by the sentiment words, including the original words and newly replaced words.

\subsection{Penalty matrix} \label{sec:penalty-maxtrix}
In neural network systems, one of the common methods for evaluating the loss functions is cross entropy \citep{bishop1995neural}. Generally, a message sample is labeled with a 3-dimensional vector $y$. Each dimension respectively represents a value in (\textit{positive}, \textit{negative}, \textit{neutral}). For example, if a message is labeled as \textit{negative}, the corresponding $y$ vector of this message is (0, 1, 0). After the learning process, a vector of probability distribution over labels of 3-dimensional $\hat{y}$ is generated, corresponding to the learning outcome of the system. The loss function is then calculated by the cross entropy formula as follows:
$$H(y, \hat{y})=-\sum y_i ln(\hat{y})$$

However, unlike standard classification task, the importance of each label in sentiment analysis is different. Generally, in this domain, the data is unbalanced. That is, the number of neutral messages is very large, as compared to other labels. Therefore, if a message is classified as \textit{neutral}, the probability that it is a misclassified case is lower than the case it is classified as \textit{positive/negative}. Moreover, \textit{positive} and \textit{negative} messages are two distinctly opposite cases. Thus, the error punished when a message expected as \textit{neutral}, is misclassified as \textit{positive}, should be less than that of the case where a negative-expected message is misclassified as \textit{positive}. The loss function is calculated by the default cross entropy function does not reflect those issues. Thus, in \citep{vo2017combination} we introduced a custom loss function, known as \textit{weighted cross entropy} in which the cross entropy loss is multiplied by a corresponding penalty weight specified in a penalty matrix.
\begin{table}[h]
    \centering
    \tbl{The penalty matrix}
    {\begin{tabular}{cccc} \toprule
        Predicted/True & Positive & Negative & Neutral \\ \midrule
        Positive & 1 & 2.5 & 2 \\
        Negative & 2.5 & 1 & 2 \\
        Neutral & 1.5 & 1.5 & 1 \\ \bottomrule
    \end{tabular}}
    \label{table:penety-matrix}
\end{table}

According to the penalty matrix in Table \ref{table:penety-matrix}, one can observe that if a message is expected to be \textit{negative} but is predicted as \textit{positive} or vice versa, the corresponding penalty weight is 2.5. Meanwhile, for the case that a message is expected to be \textit{positive} or \textit{negative} and predicted as \textit{neutral}, the penalty weight is 1.5. In other words, the former case is considered more serious than the latter. Also, if a message is expected to be \textit{neutral} and predicted as \textit{positive} or \textit{negative}, the penalty weight is 2. It is obvious that if the prediction and the expectation match to each other, the loss is not weighted as the penalty weight value is 1 (i.e the loss function is minimized in this case).

\paragraph*{Example 1.} If y is [0, 1, 0] (\textit{negative}), $\hat{y}$ = [0.2, 0.3, 0.5] (\textit{neutral}), then the default cross entropy results in 1.204, while the result of weighted cross entropy is 1.806.

\paragraph*{Example 2.} If y is [1, 0, 0] (\textit{positive}), $\hat{y}$ = [0.2, 0.7, 0.1] (\textit{negative}), the default cross entropy also results in 1.609, while the result of weighted cross entropy is 4.023.
\\
\\
Example 1 and Example 2 show that the weighted cross entropy function gives different loss values to different misclassification cases. Currently, we develop our penalty matrix based on observable intuition. However, in the future, we can rely on the distribution of data to construct this penalty matrix.

\section{Syntactic rules for handling short messages} \label{sec:rules}
As discussed, linguistic rules are very useful to handle short messages. In this section, we present the rules used in our research. These rules are suggested by some linguistic experts from the YouNet Media company \footnote{http://www.younetmedia.com} when manually working with real data. Table \ref{tab:rules} presents four basic rules that capture short messages along with examples.

\begin{table}[h]
    \centering
    \caption{Four basic rules for short messages}
    \begin{tabular}{| p{0.03\textwidth} | p{0.15\textwidth} | p{0.4\textwidth} | p{0.3\textwidth} |}
    \hline
\textbf{\#} & \textbf{Name} & \textbf{Description} & \textbf{Example} \\ \hline
1 & Directly simple rule & This rule contains some \textit{\textbf{entities}} followed by \textit{\underline{sentiment words}}; or vice versa, sentiment words followed by entities. & 
\begin{itemize}
    \item \textit{\textbf{iPhone X}} and \textit{\textbf{Samsung S9}} are \textit{\underline{powerful}} smartphone
    \item \textit{\underline{Unhealthy}} \textit{\textbf{foods}}.
\end{itemize} \\ \hline
2 & Directly comparable rule & This rule compares \textit{\textbf{entities}} with others or indicates some \textit{entities} has more or less of a quality. & 
\begin{itemize}
    \item \textit{\textbf{Electric cars}} are \underline{more expensive} than \textit{\textbf{fuel cars}}.
    \item \textit{\textbf{Fresh fruits}} are \underline{better} :).
\end{itemize} \\ \hline
3 & Product information question rule & This rule is the questions about the information such as price, release date of some product entities. It indicates the interests of the user over a product. & 
\begin{itemize}
    \item How much are these shoes?
    \item When will the new Bphone launch in Vietnam?
    \item Is this laptop available in offline stores?
\end{itemize} \\ \hline
4 & Heuristics & This rule is for messages that are not captured by the above rules, but the sentiment orientation can be inferred by heuristics. &
\begin{itemize}
    \item I am going to the supermarket
    \item Wonderful :D !
\end{itemize} \\ \hline
    \end{tabular}
    
    \label{tab:rules}
\end{table}

\section{The proposed deep architecture} \label{sec:proposed-approach}
In this section, we recall our previous work \citep{vo2017combination} first. The architecture presented in this work, now namely SALT (\textit{Sentiment Analysis for variable-Length Text}), was our first attempt to combine domain knowledge with deep learning to perform sentiment analysis in a general document. Then, we present our new architecture, known as SAINT (\textit{Sentiment Analysis for short and Informal Text}), another improved version researched in this study, aiming at short and informal textual messages.

\subsection{SALT: Combination of domain knowledge with deep learning}
\begin{figure}[h]
    \centering
    \includegraphics[scale=0.33]{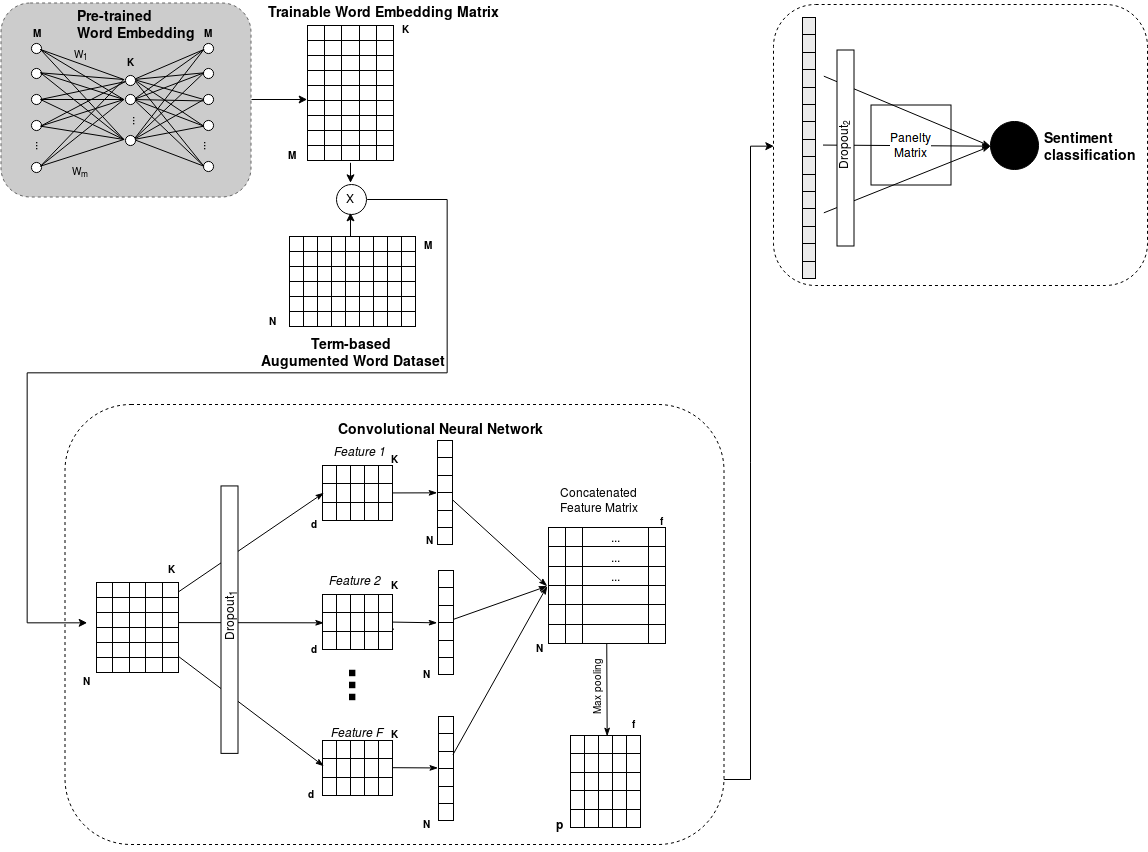}
    \caption{Previous architecture for sentiment analysis of variable-length messages}
    \label{fig:deeplearning-domain-knowledge}
\end{figure}

Figure \ref{fig:deeplearning-domain-knowledge} presents an overview of our previous work \citep{vo2017combination} of using deep architecture for sentiment analysis on variable-length messages. As previously discussed, it is referred as SALT in this paper. The SALT system includes the following modules.

\paragraph*{Word Embedding Module.} Using Skip-gram model, this is a three-layer word embedding (WE) neural network which maps words to the target words, to learn weights which act as word vector representations. The first layer consists of $M$ nodes where $M$ is the number of words in the dictionary. Each word $w$ is fed to the input layer as a one-hot vector. The hidden layer consists of $K$ neurons, where $K<<M$. The last layer also has $M$ nodes which is a softmax layer. The target words, which are one-hot vectors, are the words that appear around the input word in a context window. The error between output and target is propagated back to re-adjust the weights.

After WE is trained, the weights $w_{ij}$ on the connections from the $i$-th nodes of input layer to $j$-th nodes of hidden layer form the initial values of the $W_{M \times K}$ matrix. This embedding matrix continues to be fined-tune for sentiment task.

\paragraph*{Input Module.} This is a set of collected text, each of which was previously labeled as {\textit{positive}, \textit{neutral}, or \textit{negative}} over an object. Originally, a document of $N$ words is represented as a matrix $D_{N \times M}$, in which the $i$-th row is the one-hot vector of the $i$-th word of the document. When performing the matrix multiplication $D \times W$, we get an embedded matrix $E_{N \times K}$ of the document. The matrix $E$ is then used as input for the next \textit{Convolutional Neural Network} module.

\paragraph*{Convolutional Neural Network Module.} At this stage, the convolution is performed between the matrix $E$ with a \textit{kernel} as a $F_{d \times K}$ matrix. The meaning of the matrix $F$ is to extract an \textit{abstract feature} based on a hidden analysis of a \textit{d-gram} from the original text. There are $f$ matrices of $F_{d \times K}$ used to learn $f$ \textit{abstract features}. As the convolution of two matrices $E$ and $F$ results in a column matrix of $N \times 1$, we eventually obtain the convoluted matrix $C_{N \times f}$ by combining $f$ column matrices together.

In the next step, the matrix $C$ is \textit{pooled} by a pooling window of $q \times f$. The significance of this process is to retain the most important \textit{d-gram} feature from q consecutive \textit{d-gram}. Finally, we obtained the matrix $Q_{p \times f}$ where $p = \frac{N}{q}$.

At this point, the matrix Q is flattened and taken through a final fully connected layer to output the final result of classification (\textit{positive}, \textit{neutral}, \textit{negative}). Error from the final result is propagated back to the beginning layer of the Word Embedding module to continue the training process.

\paragraph*{The Dropout factors.} As discussed, in order to avoid the overfitting, we use the Dropout technique \citep{srivastava2014dropout}. The application of the Dropout can be used for all systems with backpropagation learning. We used two Dropout factors $p_1$, $p_2$ respectively for the Convolutional Neural Network and final fully connected layers.
\\
\\
\indent In this deep architecture, we integrated the domain knowledge in the following ways.

\begin{itemize}

\item We use data augmentation technique to enrich the training data based on the sentiment scores learned from domain knowledge.
\item We use the penalty matrix to make the loss function better adapt to the sentiment classification context.

\end{itemize}

\subsection{SAINT: An upgraded architecture for handling short and informal messages}
\begin{figure}
    \centering
    \includegraphics[scale=0.4]{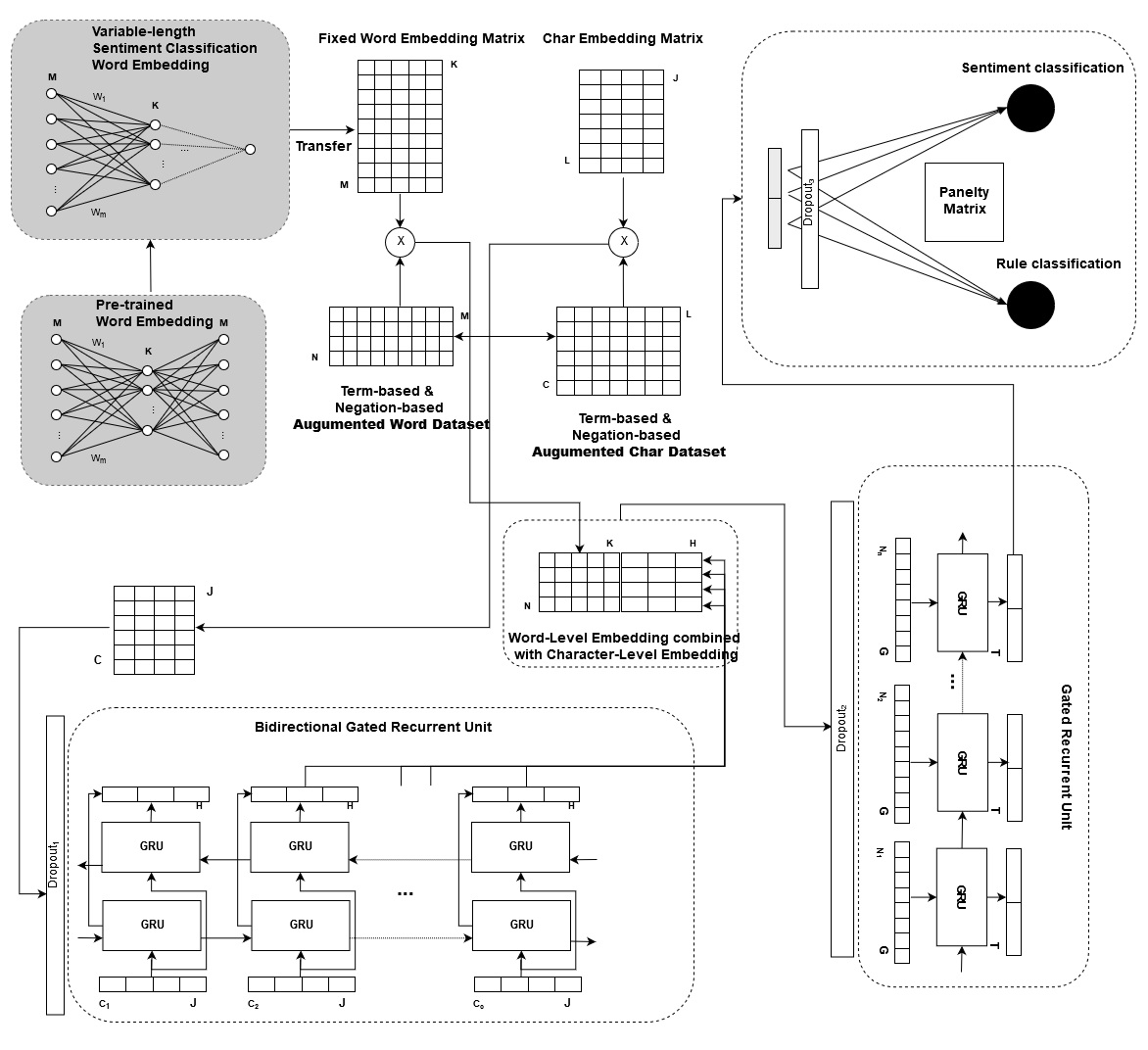}
    \caption{New architecture for handling sentiment analysis of short and informal messages}
    \label{fig:new_architecture}
\end{figure}

The previous architecture has been evolved into a new architecture, SAINT, as presented in Figure \ref{fig:new_architecture}. This time, our purpose is to effectively handle short and informal text. The new features introduced in SAINT are as follows.

\subsubsection{Negation-based augmentation} \label{sec:negation-augmentation}
An apparent contextual sentiment modifier is the negation. In a negated context, sentiment words change their polarity and with the short text, the scope of negation commonly captures the whole context. Therefore, negations are generated by adding negation words before messages or terms and reverse the scores. Beside term-based regular data augmentation, we generate augmented negated training samples as examples illustrated in Table 4. This enhancement is just a special case of the technique of data augmentation previous discussed. Nevertheless, we found that it is very useful to handle short messages.
\begin{table}[h]
    \centering
    \caption{Examples of negative-based training data augmentation}
    \begin{tabular}{|c|l|c|}
        \hline
         \# & Training data & Label \\ \hline
         1 & Bad & Negative \\ \hline
         2 & Not Bad & Positive \\ \hline
         3 & This network is stable & Positive \\ \hline
         4 & This network is hardly stable & Negative \\ \hline
    \end{tabular}
    
    \label{tab:negation_augmentation}
\end{table}

\subsubsection{Transfer learning for word embeddings} \label{sec:transfer-learning}
Intuitively, for word embedding task, we need a \textit{training corpus}. However, we found that if we develop a corpus consisting of the short message only, the performance of the classification task is not quite good, especially when we continue to train the word embeddings with the sentiment task. It is because short messages convey less context information. Thus, when used as the training data, the training process also hardly learns from the document contexts.

\begin{figure}
    \centering
    \includegraphics[scale=0.5]{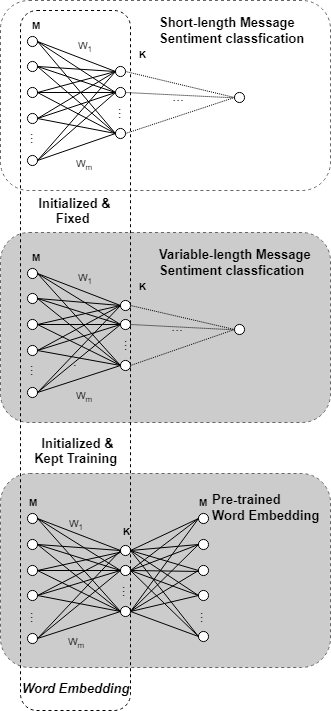}
    \caption{Transfer learning for word embeddings}
    \label{fig:transfered-we}
\end{figure}

We handle this problem by using transfer learning. Figure \ref{fig:transfered-we} depicts the process of transferring the word embedding weights through tasks. In the SALT, we already applied \textit{variable-length sentiment classification task}. In this task, the word embeddings were already trained twice. Firstly, it is learned by Skip-gram model in an unsupervised fashion. Secondly, the pre-trained word embeddings continue to be trained by the sentiment task where the backpropagation process is performed back to the embedding layer to discriminate strongly the polarity meaning of sentiment words. These are \textit{stable trained word embeddings on a large corpus}. Then, for the short-length sentiment task, we use the stable word embeddings as fixed word representations. In other words, the backpropagation process is not applied to the embedding layer.

\subsubsection{Combination of word-level embeddings and character-level embeddings} \label{sec:combination-we-cl}
\begin{figure}
    \centering
    \includegraphics[scale=0.5]{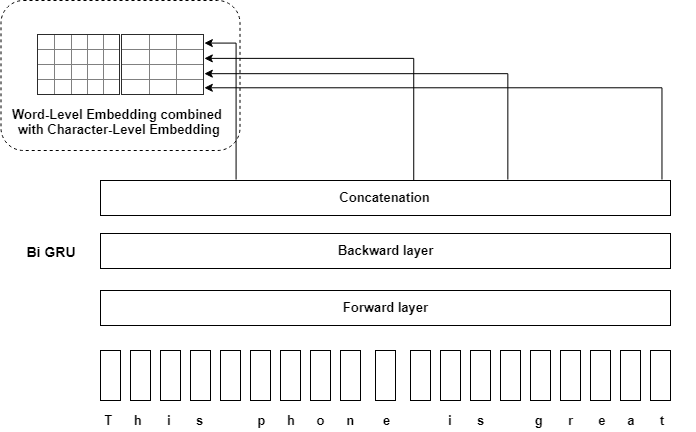}
    \caption{Combining word-level and character-level representations for the sentence "This phone is great"}
    \label{fig:combined-we}
\end{figure}

Enriching word vectors with character information is necessary to create the representations which take into account the syntactic issues. As depicted in Figure \ref{fig:combined-we} is the combination of word-level embeddings and character-level embeddings. We generate character-level word vectors by using Bi-GRUs. Its input is a sequence of vectors corresponding to characters of the input message. These character vectors are distributed representations of Vietnamese characters and learned during the training phase. The outputs of Bi-GRUs are concatenations of two states from forward and backward layers. We run Bi-GRUs on the whole message and extract the outputs at the end-character position of each word. We incorporate this model into the word-level model to get the combined representations of the inputs. 

Besides, using Bi-GRUs for representing character-level embedding captures the order information in how they account for context. Mapping of words into their concepts is an important task in understanding natural language. For instance, we can see the context of the word ``big'' is positive when it is used with words about fruit like ``orange'' or ``tree''. But the same word ``big'' has the context of negative when it is used with technological terms like ``Samsung'' and ``iPhone''.

\begin{figure}[h]
    \centering
    \subfloat[Syntactic variations of the word ``like'' in Vietnamese] {
        \resizebox*{0.45\textwidth}{!}{
        \includegraphics[scale=0.03]{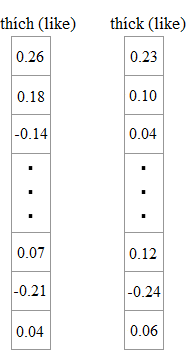}} \label{fig:character_level-a}
    }
    \hspace{5pt}
    \subfloat[The word ``big'' in context about smartphone and in context about fruit] {
        \resizebox*{0.45\textwidth}{!}{
        \includegraphics{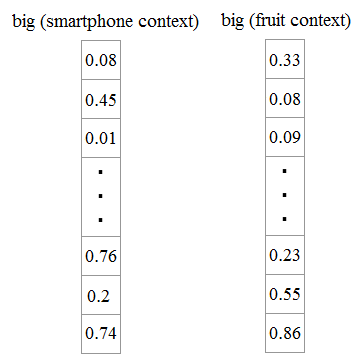}
        } \label{fig:character_level-b}
    }
    \caption{Character-level word vectors representing relationships between words}
    \label{fig:character_level}
\end{figure}

As shown in Figure \ref{fig:character_level-a}, two embedded character-level word vectors for the terms ``thích'' and ``thíck'', which both mean ``like'' in English and are the syntactic variations in Vietnamese. When trained at the word level, it is not easy to capture such syntactic similarity. However, at the character level, the cosine similarity between them is around 0.85, which reflects almost the same orientation of the two vectors. In Figure \ref{fig:character_level-b}, we illustrate the case that two embedded character-level vectors for the term ``big'' in the context of "fruit" and "smartphone" represented in the new model. In the old architecture, they are the same word. However, the new architecture learns them as two different vectors where the cosine similarity is approximately 0.6, which better reflects the context they appear in. Needless to say, those terms are significant to infer user opinions in text documents.

\subsubsection{Multi-task learning with Rules} \label{sec:multitasking}
In Section III, we already discussed the syntactic rules used to handle short messages. To integrate such rules into our deep architecture, we apply \textit{multi-task learning} by training two tasks simultaneously, known as \textit{sentiment detection} and \textit{rule classification}, to mutually enhance the performance of each task. By doing so, what is learned by rule task can help sentiment task to be learned better. As shown in Table \ref{tab:ex_short_message} are examples of messages for applying multi-task learning. Thus, apart from the prediction of sentiment results, a message sample is also additionally labeled by a rule with a 4-dimensional vector y. Each dimension respectively represents a value in {\textit{Directly Simple Rule, Directly Comparable Rule, Question Rule and Heuristic Rule}}. For example, if a message is labeled as matching with \textit{Directly Simple Rule}, the corresponding y vector of this message is (1, 0, 0, 0).

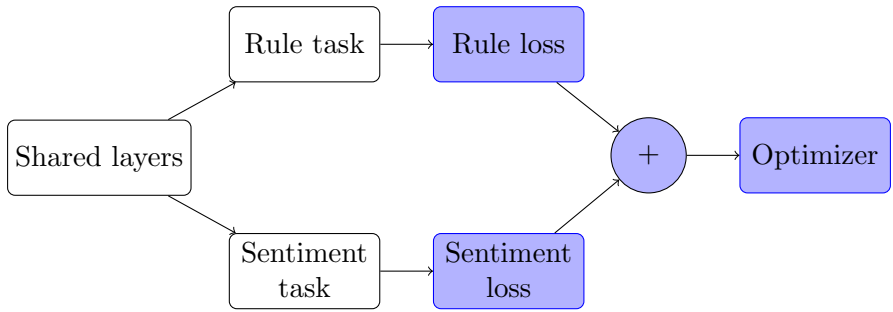
\begin{figure}
    \centering
    \begin{tikzpicture}
        [inner sep=1mm,
        node distance=0.7 cm,
        text depth=.25ex,
        inner_style/.style={rectangle, draw=black, minimum height=1cm, minimum width=2cm, rounded corners=1mm, align=center},
        outer_style/.style={rectangle, draw=blue, fill=blue!30, minimum height=1cm, minimum width=2cm, rounded corners=1mm, align=center}
        ]
        \node[inner_style]  (shared_layer) {Shared layers};
        \node[inner_style]  (rule_task)     [above right=of shared_layer]  {Rule task};
        \node[inner_style]  (sentiment_task)     [below right=of shared_layer]  {Sentiment\\task};
        \node[outer_style]  (rule_loss)     [right=of rule_task]  {Rule loss};
        \node[outer_style]  (sentiment_loss)     [right=of sentiment_task] {Sentiment\\loss};
        \node[outer_style, shape=circle, draw=black, yshift=1.8mm, minimum size=10mm]  (plus)     [above right=of sentiment_loss]  {+};
        \node[outer_style]  (optimizer)     [right=of plus]  {Optimizer};
        \draw[->] (shared_layer) -- (rule_task);
        \draw[->] (shared_layer) -- (sentiment_task);
        \draw[->] (rule_task) -- (rule_loss);
        \draw[->] (sentiment_task) -- (sentiment_loss);
        \draw[->] (sentiment_loss) -- (plus);
        \draw[->] (rule_loss) -- (plus);
        \draw[->] (plus) -- (optimizer);
    \end{tikzpicture}    
    \caption{A network that trains jointly sentiment and rule classification task}
    \label{fig:sentiment_rule}
\end{figure}

This network that performs two tasks shares the parts from the embedding layer to the GRU layer that runs through the combined word representations, while the fully-connected layer, which receives the last state of GRUs as the input, is specific to each individual task. While training, the parameters of the sentiment task do not change to the error in rule task, but the parameters of the shared layers change with both tasks. Figure \ref{fig:sentiment_rule} is a diagram that shows the network that trains tasks jointly in which the loss functions of the individual tasks is added up and optimized on that using the Gradient Descent \citep{ruder2016overview} algorithm.

\begin{table}[h]
    \centering
    \caption{Examples of rules and sentiments of short messages}
    \begin{tabular}{|c|p{0.27\textwidth}|p{0.27\textwidth}|p{0.27\textwidth}|}
    \hline
    No. & Samples & Rule Label - Encoded & Sentiment Label - Encoded \\ \hline
    1 & This phone is excellent ! & Directly simple - (1, 0, 0, 0) & Positive - (1, 0, 0) \\ \hline
    2 & A tablet and B tablet are terrible & Directly simple - (1, 0, 0, 0) & Negative - (0, 1, 0) \\ \hline
    3 & Canned beer is fresher than bottled & Directly comparable - (0, 1, 0, 0) & Positive - (1, 0, 0) \\ \hline
    4 & How much does Ipad cost ? & Question - (0, 0, 1, 0) & Neutral - (0, 0, 1) \\ \hline
    5 & Today is Wednesday & Heuristics - (0, 0, 0, 1) & Neutral - (0, 0, 1) \\ \hline
    6 & So sad :( & Heuristics - (0, 0, 0, 1) & Negative - (0, 1, 0) \\ \hline
    \end{tabular}
    
    \label{tab:ex_short_message}
\end{table}

\subsubsection{\textcolor{black}{Dropout regularization}} \textcolor{black}{It is also worth mentioning that besides the fully connected layer, all the GRU related layers deployed in our architecture are equipped with \textit{dropout} technique to prevent overfitting on our neural network systems. The dropout is applied to the input of each layer. In the experiment section, we also state the value of dropout factors chosen in our architecture.}

\section{Experiments} \label{sec:experiments}
\subsection{Experiment setting}
We have implemented a learning system based on the proposed enhanced deep architecture and conducted experiments to perform sentiment detection. Our dataset consists of 500,000 short and informal documents whose lengths are less than 45 characters, collected from the social media channels of Facebook, YouTube, Instagram, forum, e-newspapers and blogs in Vietnam. The dataset is split into training set, validation test and testing set with the ratio of 60:20:20. The dataset was labeled and provided by YouNet Media, a company specializing in online data analysis. The company also provides a \textit{sentiment dictionary} that includes positive and negative terms along with their scores.

Initially, the data were represented as one-hot vectors consisting of 65000 dimensions. After performing the word embedding technique, these vectors were reduced to 320 dimensions. Each message is truncated or padded to be the same word length of 10. Character-level word vectors are concatenations of two states from forward and backward layers of Bi-GRUs which uses 100-unit cells running through character sequence. We combine word-level embeddings with character-level embeddings to create the matrix having a size of $10 \times 420$ representing the embedded text. Then, we apply GRUs using 200-unit cells sequentially on each word in a sequence of 10 combined word vectors and get the last state as the input to the fully connected layer. All dropout factors are set at 0.7.

\subsection{Experimenting models}
In our experiment, we developed various models, as presented in Table \ref{tab:experiments_models}. Each model deploys various enhancement techniques as follows.
\begin{itemize}
    \item \textit{Regular Argumentation}: It implies the data argumentation technique using domain knowledge which we published in \citep{vo2017combination}.
    \item \textit{Penalty matrix}: This technique is also published in \citep{vo2017combination}, which is recalled in Section \ref{sec:penalty-maxtrix}
    \item \textit{Negation Augmentation}: It refers to the negation-based argumenation technique discussed in Section \ref{sec:negation-augmentation}
    \item \textit{Transferred WE}: It refers to the transfer learning technique discussed in Section \ref{sec:transfer-learning}
    \item \textit{Combined WE-CE}: It refers to the enhancement technique discussed in Section \ref{sec:combination-we-cl}
    \item \textit{Multi-task}: It refers to the multi-task learning technique discussed in Section \ref{sec:multitasking}
\end{itemize}

Regarding the core classification techniques, we employ the well-known techniques of Support Vector Machine (SVM), CNNs and GRUs. CNNs are considered as the baseline method for our works (Model \#2), as we mainly rely on DNN approaches. Note that Model \#4 is corresponding to our previous work in \citep{vo2017combination} and the Model \#9 is our eventual model of SAINT. \textcolor{black}{Moreover, we also compare our approach with two other aforementioned models of \citep{Santos2014} and \citep{wang2016combination}, which are quite well-known in this area. Those two additional models are referred as CharSCNN and CNN-RNN, respectively}.

\begin{table}[h]
    \centering
    \caption{Classification models using various approaches}
    \begin{tabular}{|c|c|p{0.45\textwidth}|c|}
        \hline
        \textbf{\#} & \textbf{Model} & \textbf{Enhancement} & \textbf{Notes} \\ \hline
        1 & SVM & & \\ \hline
        2 & CNN & & Baseline \\ \hline
        3 & CNN & Regular Augmentation & \\ \hline
        4 & CNN & Regular Augmentation, Penalty Matrix & SALT \\ \hline
        5 & CNN & Regular Augmentation, Negation Augmentation Penalty Matrix & \\ \hline
        6 & CNN & Regular Augmentation, Negation Augmentation Penalty Matrix, Transferred WE & \\ \hline
        7 & CNN & Regular Augmentation, Negation Augmentation Penalty Matrix, Transferred WE, Combined WE-CE & \\ \hline
        8 & GRU & Regular Augmentation, Negation Augmentation Penalty Matrix, Transferred WE, Combined WE-CE & \\ \hline
        9 & GRU & Regular Augmentation, Negation Augmentation Penalty Matrix, Transferred WE, Combined WE-CE, Multi-task & SAINT \\ \hline
        \textcolor{black}{10} & \textcolor{black}{\citep{Santos2014}} & & \textcolor{black}{CharSCNN} \\ \hline
        \textcolor{black}{11} & \textcolor{black}{\citep{wang2016combination}} & & \textcolor{black}{CNN-RNN} \\ \hline
    \end{tabular}
    \label{tab:experiments_models}
\end{table}

\subsection{Experiment results}
Our experimental results are presented in Table \ref{tab:experimental_results}. We are based on the standard metrics of \textit{Precision}, \textit{Recall} and \textit{F-measure} to evaluate the performance of the experimenting models.

The first observation is that the deep learning approaches outperform the classical technique of SVM. In addition, all of the enhanced models also achieve better performance as compared to the baseline model (\#2).
 
In SALT (Model \#4), when the penalty matrix is applied, the precision is increased significantly, although the recall is slightly reduced. It implies that the penalty matrix achieves better accuracy when detecting targeted cases (e.g. detecting negative messages), but this may cause missing some appropriate cases. When the enhancement techniques of Negation Augmentation, Transferred WE and Combined WE-CE are applied in Model \#7, all metrics of Precision, Recall and F-measure are increased visibly. Further significant improvement is also observable when we replace the core technique from CNNs to GRUs (from Model \#7 to Model \#8).

Besides, when compared to other existing works of CharSCNN and CNN-RNN (Model \#10, \#11), we observe that those two models outperformed our old model of SALT. In fact, the performance of those two CNN-based models are quite similar to our CNN-based models used in the experiments. Finally, when all enhancements are combined in SAINT (Model \#9), the best performance is enjoyed.

\begin{table}[h]
    \centering
    \caption{Experimental results}
    \begin{tabular}{|c|c|c|c|}
        \hline
        \textbf{Model} & \textbf{Precision} & \textbf{Recall} & \textbf{F-measure} \\ \hline
        1 & 0.781 & 0.795 & 0.788 \\ \hline
        2 & 0.803 & 0.811 & 0.807 \\ \hline 
        3 & 0.807 & 0.820 & 0.813 \\ \hline 
        4 & 0.824 & 0.813 & 0.818 \\ \hline 
        5 & 0.832 & 0.821 & 0.826 \\ \hline 
        6 & 0.841 & 0.829 & 0.835 \\ \hline 
        7 & 0.856 & 0.844 & 0.850 \\ \hline 
        8 & 0.859 & 0.851 & 0.855 \\ \hline 
        \textbf{9} & \textbf{0.867} & \textbf{0.857} & \textbf{0.862} \\ \hline
        \textcolor{black}{10} & \textcolor{black}{0.819} & \textcolor{black}{0.843} & \textcolor{black}{0.831} \\ \hline
        \textcolor{black}{11} & \textcolor{black}{0.838} & \textcolor{black}{0.847} & \textcolor{black}{0.842} \\ \hline

    \end{tabular}
    
    \label{tab:experimental_results}
\end{table}

\section{Conclusion} \label{sec:conclusion}
In this paper, we continue extending our work in \citep{vo2017combination}, in which a method for a combination of domain knowledge and deep learning approach was introduced. In this study, we focus on specific problems of handling short and informal messages, which commonly occur on social media channels. Existing works using deep learning are currently suffering from various difficult issues when dealing with this kind of data.

To overcome those problems, we suggest some further enhancement techniques, including \textit{negation-based data augmentation}, \textit{transfer learning} for word embeddings of short messages, \textit{combination of word-level and character-level} for word embeddings and \textit{multi-task learning}. The core classification technique is also changed from CNNs to GRUs to better reflect the nature of sequence data. When experimenting with real datasets from social media channels, these enhancements make us enjoy significant improvement in performance.

In the future, we consider applying the attention-based approach of casting NLP tasks into question answering (QA) problems, which was first introduced in \citep{kumar2016ask}, to have a unified architecture to solve various NLP problems of sentiment analysis, entity recognition, intention detection and question answering. If we successfully integrate domain knowledge into such an approach, we may hopefully achieve better performance from real datasets.

\pagebreak


\bibliographystyle{agsm}
\bibliography{ref}
\end{document}